# Improving ICD coding using Chapter based Named Entities and Attentional Models

Abhijith R. Beeravolu, Mirjam Jonkman, Sami Azam , Friso De Boer

**Abstract**
Given considerable breakthroughs in natural language processing (NLP), many state-of-the-art methods were created to automate various processes. However, NLP in the clinical domain is mainly limited to curated and benchmark datasets, which may not reflect how well they can operate in real-world situations. Automatic ICD coding is one such NLP process that heavily relies on public datasets (MIMIC-III), which are imbalanced and old. Most of them have used the attention mechanism to assign weights to the features in the dataset that belong to each ICD code. The F1 scores from the existing methods show that the micro-averages are between 0.4 - 0.7, suggesting that there are still a lot of false positives. Our research allowed us to propose an approach that can improve the ICD coding and achieve better F1-scores. This approach uses chapter-based named entities and attentional models to perform two tasks: categorize discharge summaries in ICD-9 Chapters and create an attentional model using only the data that belongs to a chapter because the models don't need to see or know the outside data to identify codes within a chapter. For categorizations, we have proposed an approach using Chapter-IV for de-biasing and influencing the important entities and weights of the Chapter without relying on neural networks. This allowed us to create categorization thresholds that can fit the data perfectly and provide faulty threshold values and interpretability for the human coders to correct and validate the categorizations. Once the categorizations are validated, we have selected 3 frequent and 3 non-frequent codes that belong to Chapter-IV to create the attentional models using two architectures: Bidirectional-Gated Recurrent Units (GRUs) + Attention and Transformer with Multi-head Attention. The selected codes reflect most of the codes in Chapter-IV. Averaging the Micro-F1 scores from all six models for the two architectures gives 0.79 and 0.81, showcasing significant improvements in the performance of ICD coding.

## I. INTRODUCTION

Clinical narratives serve as the primary mode of communication in the healthcare industry, offering a unique account of a patient's history and assessments [1] that can be used in both the clinical context and research. However, the majority of valuable information is in unstructured/semi-structured text [1][2] that is large, unrestrictive, and ambiguous [2]. Sometimes using human language is not precise enough to communicate information, especially in healthcare; it is sometimes difficult to understand or interpret another clinician's diagnosis summary if the language consists of ambiguities, metaphors, and colloquial expressions [2]. To tackle this problem, several standard computer-readable vocabularies (ICD, RxNorm, SNOWMED, etc.,) have been created for healthcare systems so that clinical and administrative information can be communicated effectively. For decades, healthcare providers have relied on manual 'medical coders' for coding diagnosis or treatment information in a clinical text based on well-maintained vocabularies. With significant advances in natural language processing (NLP), researchers have proposed methods to automate/semi-automate the processes, showcasing state-of-the-art performances. However, performance evaluation of such approaches has been limited to curated and clean benchmark datasets that may not properly reflect how robustly these systems can operate in real-world situations [3][32]. And one key obstacle in the clinical domain is data accessibility, mainly due to ethical constraints [32] on sharing documents with personal information and strict governance regulations to de-identify the electronic health records (EHR) data [4].

International Classification of Diseases (ICD) is one such standardized vocabulary that relied on manual code assigners for a long time. And existing methods for automating the ICD code assignment using the clinical text have heavily relied on the benchmark (MIMIC-III) [5] and curated datasets. With MIMIC-III, there are only ~9000 ICD-9 codes in total, covering 58,976 unique admissions with a mean of 11 ICD codes per admission (min: 1, 25%: 6, 50%: 9, 75%: 15, max: 39). Various state-of-the-art architectures and methods [6] – [15] for ICD coding were proposed using MIMIC-III. Most of them have used the attention mechanism to assign weights to words/text snippets in the dataset associated with each ICD code. Along with the semantic features from the clinical notes, researchers have also integrated knowledge bases that are created using original ICD code titles and associated synonymous terminology to improve code classification [16] – [19]. The results available from these methods (on MIMIC-III) showcase that micro-average F-scores are in the range of 0.4 to 0.7, with the area under the curve (AUC) close to 1, suggesting that there are still a lot of incorrect predictions *(see Table. 1)*. This could be due to various factors such as:

*1) Class imbalance*

Most of the ICD codes in MIMIC-III consist of imbalanced data with large deviations between *normal* and *anomalous* classes, and some ICD codes only have a few samples or even no samples for training the model, which results in poor prediction accuracy in rare labels and even providing unreliable results for frequent codes due to class imbalance.

*2) High false-positives from the models (or feature weights)*

To demonstrate this point, we have selected the ICD codes (280 - 289) from Chapter – IV *(Diseases of the Blood and Blood-Forming Organs (ICD-9))*. Most of the diseases and conditions (ICD-9 codes) that belong to Chapter-IV in MIMIC-III are not part of the primary diagnosis (i.e., code is not present at index 0 or 1 (mostly are primary) in the ICD code list that belongs to a patient/discharge summary). In MIMIC-III, there are 19,006 (*total = 52,726*) unique discharge summaries that belong to Chapter-IV. Reducing the original ICD code list of a patient to 'list of 3' and 'list of 5' (i.e., from the head) allowed us to identify whether Chapter-IV codes are present in them or not. Out of 19,006 discharge summaries, there are only 3,328 and 6,632 summaries that belong to the reduced lists 3 and 5, which suggests that codes from Chapter-IV mainly arise due to adverse reactions from prescribed drugs or complications in various anatomical regions. Counting the occurrence of each ICD code at index 0 and 1 showcases that Chapter-IV codes are mainly associated with the chapters: *diseases of the circulatory system, infectious and parasitic diseases, and diseases of the digestive system* (at Index 0); *diseases of the genitourinary system, diseases of the circulatory system, and diseases of the respiratory system* (at Index 1). These associations can cause the models (or feature weights) to provide wrong codes for many discharge summaries.

We have implemented two architectures: Bidirectional-Gated Recurrent Units (GRUs) with Attention and a Transformer with Multi-head Attention, to fit the *normal* class against the *anomalous* class, that is, to extract feature weights about Chapter-IV, all the discharge summaries in MIMIC-III that belong to that Chapter (19,006) are labelled as 'normal' and others in the dataset as 'anomalous' (33,720). This allows us to see how the models (or feature weights) distinguish between the ICD codes of Chapter – IV (280-289) and the Rest of the codes. All the summaries are processed to remove unnecessary information (symbols, numbers, whitespaces, etc.) that might degrade the performance; and create a lower-case string (summary) with a continuous sequence of words from head to tail. To train the models, we used a maximum sequence length (M) of 3000 as the input length because it covers ~95% of the data and is two standard deviations from the average (~1500). And all the other parameters are selected based on the trial and error method. When tested using a test-set {1: 2,773, 0: 4,993}, the models gave TPR of 75% (1) and TNR of 68% (0) for the transformer model in Table. 1. Even with bias, the transformer model at epoch 20 is classifying the *anomalous* class as *normal*, producing many false positives.

**TABLE I**
MODEL: CHAPTER – IV CODES (280-289) VS REST OF THE CODES
(M: MAX. SEQUENCE LENGTH, E: EMBEDDING DIM, G: GRU DIM, D: DROPOUT, B: BATCH SIZE, EP: EPOCHS, N: NUMBER OF HEADS, DF: FEED FORWARD UPWARD PROJECTION SIZE (DFF), TPR: TRUE-POSITIVE RATE, F1: F1-SCORE, PRECISION, AND RECALL)

| Algorithm | Train 1: 16,273 0: 28,727 | Test (1: 2,773; 0: 4,993) | | | |
|---|---|---|---|---|---|
| | | TPR TNR | F1 | P | R |
| BiGRUs + Attention | M=3000 E = 100 G = 1024 D = 0.2 B = 64 | Ep = 5 1: 0.66 0: 0.82 Ep = 3 1: 0.77 0: 0.80 | 0.67 0.82 0.72 0.83 | 0.67 0.82 0.68 0.86 | 0.66 0.82 0.77 0.80 |
| Transformer + Multi-head Attention | M=1500 E = 100 N = 4 Df = 128 D = 0.2 B = 64 | Ep = 20 1: 0.75 0: 0.68 Ep = 3 1: 0.69 0: 0.86 | 0.64 0.75 0.71 0.85 | 0.56 0.83 0.73 0.84 | 0.75 0.68 0.69 0.86 |

To increase the correct predictions in the transformer model, it seems we should sacrifice a lot of true-negatives (18%) (i.e., more wrong classifications). Assigning incorrect ICD codes for a patient diagnosis or treatment can substantially increase the costs for patients and insurance providers [20][21].

*3) ICD codes with no associated words (synonyms) in some discharge summaries*

In MIMIC-III, there are discharge summaries that describe similar conditions differently. Some of these summaries don't have medical words (or features) that can be directly linked with an ICD code name or associated synonyms, as others do. This suggests that some of the codes for discharge summaries in MIMIC-III are based on various disease conditions and their associated symptoms. This can cause the models to not provide some codes for the discharge summaries during prediction because, for some discharge summaries, there are fewer 'attentional features' in the text than others.

*4) Unnecessary information in the discharge summaries*

Most of the discharge summaries in the MIMIC-III dataset are organized into different sections, such as history of present illness, past medical history, various examination reports, pertinent results, brief hospital course, medications on admission and discharge, discharge diagnosis, and others. Some of these sections are not useful during auto-ICD coding and could add noise to the models during training. For example, in the section 'past medical history,' there is past diagnosis information that has nothing to do with current admission or diagnosis. When we use past medical information with the present information to train a model, the features obtained by the model from the past section can provide codes that are not related to the present admission to discharge information. This can result in a lot of incorrect or prediction of codes for full-length discharge summaries in MIMIC-III. To tackle this problem, we have applied various regular expression patterns (*see* Section- II) to remove unnecessary sections and create a 'short-summary' that only has information related to present admission till discharge. For the Rest of the research, we have used the short summaries for training the algorithms and extracting feature weights associated with different labels (or codes).

Various researchers have proposed methods for auto-ICD coding by implementing approaches such as: using ~9,000 ICD codes (all) or ~7000 ICD codes in MIMIC-III and using the Top 100 or 50 codes in MIMIC-III [6][8][16]. Most proposed methods are evaluated using AUC scores (along with others) with specious values greater than 90%. However, the f1-scores (0.4-0.7) for these methods don't have specious values and tell a different story about the overall performance. If the extracted features cannot differentiate *Chapter - IV Codes vs Rest* accurately, applying the attention mechanism could not produce good attentional models/weights with fewer incorrect predictions.

The current version of ICD (ICD-11) has over 55,000 diagnostic codes, up from around 14,000 with ICD-10, making it difficult for researchers to propose methods (using MIMIC-III or others) that won't result in high redundancies when capturing the features for all the ICD codes separately. And the existing methods that are proposed using ~9000 codes have assigned attentional weights for all the codes together from the dataset (52,726). This can cause the codes to have similar features (with different weights) in their attentional models, resulting in many incorrect predictions. Therefore, rather than creating the attentional models directly from the entire dataset, this research would like to create and separate the attentional models according to ICD chapters, as shown in Fig. 2. To create the attentional weights for codes in a chapter, only data (in MIMIC-III) that belongs to that Chapter will be used, because the algorithm during training does not need to see the data from chapters that are outside, which is not the case in most of the existing methods.

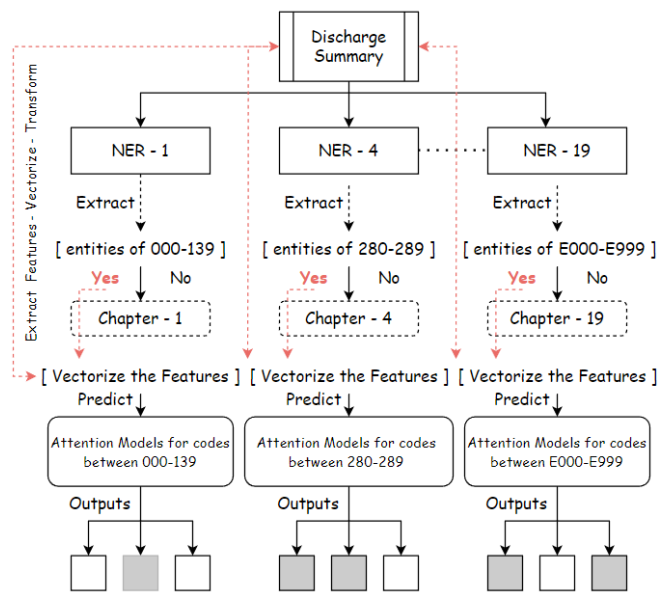

*Figure 2: Using separate Named Entity Recognition (NER) and Attention Models for the ICD Chapters (ICD-9).*

In recent years, many deep neural network models have achieved state-of-the-art performance gains in various natural language processing (NLP) tasks [20]. However, these gains mostly relied on the availability of large amounts of labelled examples, without which state-of-the-art performance is rarely achievable [21]. This is especially inconvenient for many NLP fields where labelled examples are scarce, such as medical texts. Our research believes combining named entity recognition (NER) with deep learning (or attention mechanism), as shown in Fig. 2, can improve the performance of auto-ICD coding.

Initially, the NER models are used to categorize the short-summaries according to their chapters, and then the attentional models that belong to the chapters are used to provide codes for the categorized summaries. The NER models can provide interpretability by listing the medical entities (diseases, symptoms, etc.,) that are present in a discharge summary. These entities can be used to validate or correct the categorization. And the attentional models for the respective chapters are created using only the data that belongs to those chapters; that is, to create an attentional model (or feature weights) for code 285.9, we should create a dataset based on *one-vs-rest* approach (285.9 vs. Rest in Chapter – IV). This allows us to extract better feature weights for the ICD codes than before and improve the overall performance. Section – II describes in detail the methods used in this research, from processing the discharge summaries to the prediction of the codes.

## II. METHODS

### A. Data Creation

The dataset used for this study is MIMIC-III, which is a freely available database that consists of health-related information associated with distinct hospital admissions. Most of the existing methods for ICD coding have used discharge summaries for performing their experiments. To create the datasets for this research, two major tables from MIMIC-III were selected: *Diagnosis_ICD & Noteevents.*

#### 1) Dataset

The *Diagnosis_ICD* dataset consists of 58,976 unique HADM_IDs with a mean of 11 ICD codes per admission. In MIMIC-III, each ICD code of a patient is separated by a 'row'. Therefore, the ICD codes are grouped together based on their HADM_IDs.

And the *Noteevents* dataset consists of medical texts that belong to 15 categories (Discharge Summaries: 59,652). The datasets *Diagnosis_ICD and Noteevents* are grouped and merged based on their HADM_IDs, creating 52,726 discharge summaries.

#### 2) Labelling

To label the data according to the ICD codes (ex: 285.9 vs Rest in Chapter - IV), we have created an algorithm (see Fig. 5) that can create the required datasets automatically, that is, labelling 285.9 as *normal* (1) and other ICD codes between 280-289 as *anomalous* (0). Before applying the *labelling* function (Fig. 4), we used the values 2, 3, 4, 5 to extract digits from the ICD codes based on their index (see Fig. 3). This allows us to easily divide the original data according to its chapters and sub-chapters. For example, extracting the first two digits (*28*) of the ICD codes (*281.1, 285.1*) by using the value '2' allows us to easily divide the data according to its chapters (*Chapter-IV (280-289)*). Similarly, value '3' is good for the sub-chapters.

*Figure 3: Merging Discharge Summaries and ICD9_Codes based on HADM_IDs; and reducing the ICD codes based on index values {2, 3, 4, 5} to label the dataset chapters/sub-chapters) according to One-vs-Rest (OVR) approach*

```
Algorithm: Labelling data according to chapters/sub-chapters (One-vs-Rest)
1  function labelling(row);
Input:  (ICDCodeList) List of ICD codes that belong to a chapter/subchapter
        (MergedCodes) List of all the ICD codes for a HADM_ID
        (Common)     Empty list for storing common codes between
                     ICDCodeList & MergedCodes
Output: Labelled data according to one-vs.-rest approach.
5  for each code in ICDCodeList do
6      if code is present in MergedCodes then
7          append the code to Common list
8      endif
9  end for
8  if length of the Common list is >= 1 then
9      return '1'
10 elif length of Common list if == 0 then
11     return '0'
12 endif
```

*Figure 4: Pseudocode for labelling the data according to ICD codes*

#### 3) Pre-processing of Discharge Summaries

Using simple regular expression patterns (see Table. 2), we have searched (Search()) and extracted the required sections present in the discharge summaries: history of present illness, pertinent results, brief hospital course, and discharge diagnosis. Not all the discharge summaries (52,726) in MIMIC-III follow a similar structure or have similar section headings, but most of them do. Table. 2 showcases the patterns that are used and the number of matches that are found in MIMIC-III.

TABLE 2
4 REGULAR EXPRESSION PATTERNS USED FOR EXTRACTING THE REQUIRED SECTIONS AND THEIR MATCHES

| Section | Regular Expression Pattern | Matches in 52,726 Summaries |
|---|---|---|
| History of Present Illness | re.search(r'history of present illness(.*?)past medical history', sentence).group() | 46,552 |
| Pertinent Results | re.search(r'pertinent results(.*?)brief hospital', sentence).group() | 37,285 |
| Brief Hospital Course | re.search(r'hospital course(.*?)discharge medications', sentence).group() | 42,409 |
| Discharge Diagnosis | re.search(r'discharge diagnosis(.*?)discharge', sentence).group() | 39,871 |

We want to create data that covers all the four sections. To do so, we have merged all the matches found for the sections, creating a dataset of 35,352 discharge summaries (or short-summaries). Finally, these short-summaries are processed (cleaned) using various regular expression patterns to remove symbols, numbers, whitespaces, newline-spaces, etc. And all the words in the summary are converted to lower-case letters so that the model won't assign separate weights for *Anemia* and *anemia*. This allowed us to create short-summaries with continuous sequence of words from beginning to end, with no breaks or numbers. These short-summaries are used for creating NER models for chapter-level prediction and attentional models for ICD code predictions. Using "short-summary/version" of the discharge summaries showcased similar or better performance than full-length summaries. Comparing the results in Table. 1 and Table. 3 suggests that, short-summaries can classify (on a test-set) the *normal* class with 75% true-positive rate when trained till 3 epochs, which is better than the full-length results at epoch 3 (67%) (see Table 1). Here, we have used a maximum sequence length (M) of 1000 and GRU dimension (LSTM units) of 1024 for training the models with short-summaries, compared to 3000 and 1024 for full-length. The M values are assigned by calculating the mean length of sequences in the dataset. And the GRU dimension is decided based on the trail and error method. For example, training the models with full-length summaries using the parameters of short-summary models in Table. 3, the true-positive and -negative rates at epoch 3 are 59% (1) and 90% (0), which is evident that the parameters should be adjusted based on the datasets.

**TABLE 3**
BI-GRUS + ATTENTION FOR SHORT SUMMARIES (CHAPTER-IV VS REST)

| Model | Train<br>1: 15,293<br>0: 20,059 | Test (1: 2,266; 0: 3,086) | | | |
|---|---|---|---|---|---|
| | | TPR | F1 | P | R |
| BiGRUs<br>+<br>Attention | M=1000<br>E = 100<br>G = 1024<br>D = 0.2<br>B = 64 | Ep = 20<br>1: 0.69<br>0: 0.73<br>Ep = 3<br>1: 0.72<br>0: 0.81 | 0.67<br>0.73<br>0.73<br>0.79 | 0.63<br>0.77<br>0.74<br>0.80 | 0.69<br>0.73<br>0.72<br>0.81 |

### B. Named Entity Categorization (NEC) Models

Machine reading (MR) is essential for unlocking valuable knowledge present in millions of medical documents of an EHR [22]. Recognizing entities such as body parts, diseases, disease conditions, and symptoms from discharge summaries can be helpful in categorizing the data according to its main chapters. Catling, F. et al. [23] in their research used text representations from gated recurrent units (GRUs) to classify the discharge summaries (MIMIC-III) into 19 chapter-level (Level-1) labels with f1-scores of 0.68 – 0.7. GRUs are particularly attractive, as they produce a similar performance to long short-term memory (LSTMs) while using a simpler design with fewer trainable parameters [24]. LSTMs and GRUs have been used to represent sequential healthcare data, such as multivariate time series, text documents, and others [25]. Various researchers have utilized them to detect the word and character-level features automatically in texts using NER and POS-tagging [26], eliminating the need for most feature engineering. However, in this research, rather than using CNN and LSTMs directly for classifying the discharge summaries into 19 Chapters, we propose to create simple and separate named entity categorization models for each Chapter (i.e, 19 Models). The NEC models created in this research do not have many underlying machine-learning components. The only component we used is a pre-trained model (*en_ner_bc5dr_md*), which is used for extracting "DISEASE" entities present in all the short summaries (35,352) (see Fig. 5). It is crucial to notice that the pre-trained model we used has F1-Score of 84.53% [27], which means, not all the medical entities present in the summaries can be identified by the pre-trained model. There will be some unnecessary entities (Ex: 'of,' 'the,' 'and,' 'pain,' 'htn,' etc.). From the entities of short-summaries in Chapter-IV (15,293), we have selected 375 of the most prevalent entities, which are then reduced to 300 after cleaning (Fig. 5).

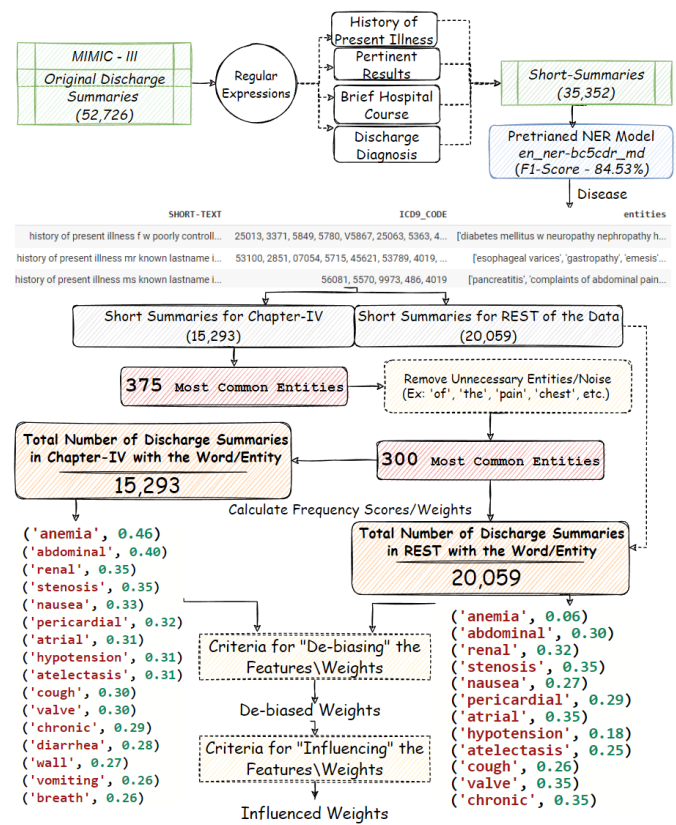

*Figure 5: Approach/Steps for Named Entity Recognition-Categorization (NER-C) Model of Chapter-IV*

Based on the 300 entities, we generated two sets of frequency scores/weights (Chapter-IV, REST) as shown in Fig. 5. According to the criteria mentioned below, the entities\weights for Chapter-IV are 'de-biased' and 'influenced' using the weights generated for the REST dataset.

#### 1) Criteria for De-biasing

After generating two sets (set1 and set2) of frequency scores/weights for the 300 entities, we have subtracted them (set1 – set2) so that the weights in set2 that are greater than or equal to set1 can be removed (see Fig. 6). For example, the entities 'stenosis' and 'chronic' has weights of {0.35, 0.35} and {0.29, 0.35} in the two sets (Fig. 6). Subtracting the set2 weight's from set1 produces values {0} and {-0.06}, which means, if the entities 'stenosis' and 'chronic' are used for categorizing the short summaries of Chapter-IV, they could be responsible for the wrong classification of summaries in the REST dataset. Therefore, our criteria for de-biasing is to remove the entities that have negative values or 0 after subtracting both sets (see Fig. 6). This reduced the total entities from 300 to 141. And the categorization results (for Chapter-IV) obtained using the de-biased weights of the 141 entities showcased better or similar performance to that of the original entities and weights (300) from Chapter-IV (set1) and the attention model in Table. 3.

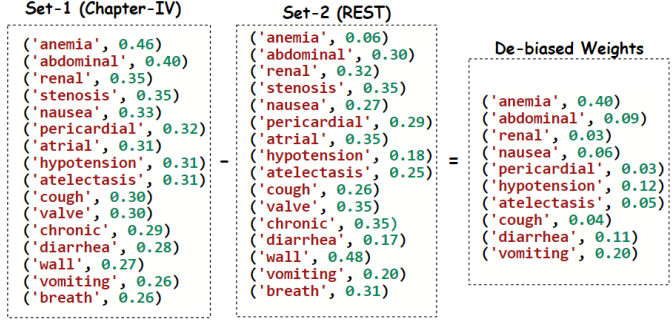

*Figure 6: Criteria for De-biasing*

*2) Criteria for Influencing*

After creating the de-biased weights, we have influenced them using two rules/criteria:
- Create a list of dictionary items closely related to Chapter-IV according to literature and some official websites. And add the new items to the de-biased weights list by assigning a weight of 0.5 for all of them (see Fig. 7).
- If the difference between the weights of entities in set1 and set2 is greater than 0.1, then the weights for those entities are doubled (see Fig. 7).

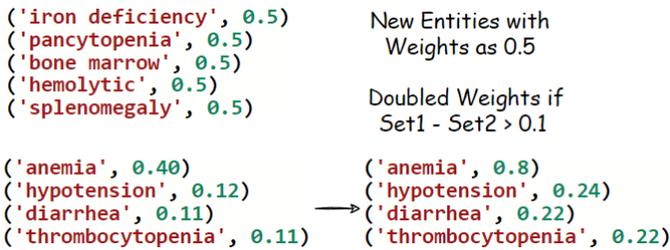

*Figure 7: Criteria for Influencing*

Currently, the dictionary list is limited to five entities only. This list can be expanded by doing more research related to diseases of blood and blood-forming organs (Chapter-IV). Adding the new entities and increasing the weights of some old ones (4) has improved the performance of categorization for Chapter-IV.

### C. Attention Models for Codes between 280-289 (IV)

In this research, we have implemented two simple network architectures that are based on the attention mechanism [28]: Bidirectional Gated Recurrent Units (Bi-GRUs) with Attention and a Transformer with Multi-Head Attention. . For the past few years, the attention mechanism has garnered a significant amount of interest, particularly in sequence tasks [29]. The term "attention" can have quite a few different meanings, depending on who you ask or where you look, but the one that applies best to this research is as follows: The attention mechanism is used to describe a weighted average of (sequence) elements, with the weights being dynamically computed based on an input query and the keys of the elements, as shown in the equation below.

$$\text{Attention}(Q, K, V) = \text{softmax}\left(\frac{QK^T}{\sqrt{d_k}}\right)V$$

Here, Q is a set of queries $Q \in R^{T \times d_k}$ for each token in an input sentence, which are used to match against a series of keys $K \in R^{T \times d_k}$ that describe values $V \in R^{T \times d_v}$, where T is the sequence length, and $d_k$ and $d_v$ are the hidden dimensionality for queries/keys and values respectively.

Most attention mechanisms differ in terms of what Queries they use, how the Key and Value vectors are defined, and what Score function is used. Sometimes using a single weighted average is not enough to attend to multiple different aspects present in a sequence, thus, allowing the extension of the attention mechanism to multiple heads, which runs through the attention mechanism several times in parallel (multi-head Attention). Using Multi-attention heads allows us to attend to the parts of an input sequence differently [28]. Specifically, given a Q, K, and V matrix, we transform them into **h** sub-queries, sub-keys, and sub-values passed through the scaled dot product attention independently. Afterward, we concatenate the heads and combine them with a final weight matrix. Mathematically, we can express this operation as:

Multihead(Q, K, V) = Concat(head$_1$, …., head$_h$)W$^O$
Where head$_i$ = Attention(QW$_i^Q$, KW$_i^K$, VW$_i^V$)

Using these architectures, we have created the attentional models for six codes between 280-289 {3: frequent, 3: non-frequent}. To create training and testing datasets for the architectures, we have used the short-summaries (15,293) that belong to Chapter-IV. These summaries are labelled based on *one-vs-rest* approach, that is, labelling the summaries for the code 285.9 as '1' and others in 280-289 as '0'. This is how we labelled the datasets for all the codes we used in this research. Creating separate attentional models for codes in Chapter-IV using only the data within Chapter-IV showcased better performance than creating the attentional models directly for all the codes.

## III. IMPLEMENTATION AND RESULTS

### A. Named Entity Categorization (NEC) Model

In this research, we are not building NEC models for all the chapters (19), rather, we are building 1 NEC model for Chapter – IV (*diseases of the blood and blood-forming organs*). This model is used for performing the experiments on the de-biasing and influencing criteria used in this research. Fig. 9 showcases how the final weights and entities (146) are used for categorizing the short-summaries into Chapter-IV.

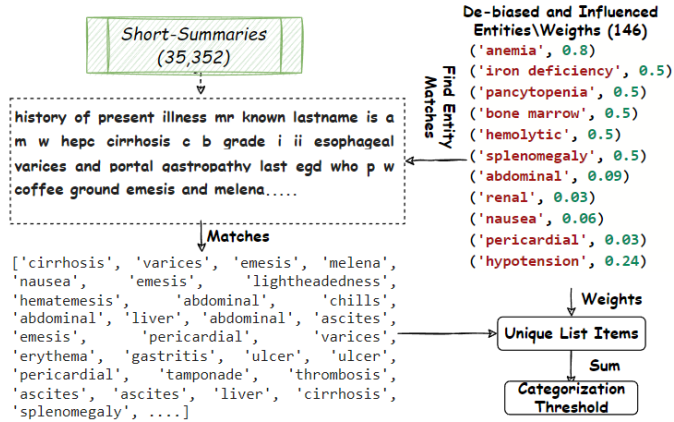

*Figure 9: Categorization of Short-Summaries into Chapter-IV*

As shown in Fig. 9, the de-biased and influenced (146) entities are used to find the matches in the short summaries. Using the weights of 146 entities, we can add the weights of all the unique items in the matched list. This sum is used for creating the categorization thresholds for Chapter-IV and REST (see Table. 4).

**TABLE 4**
CATEGORIZATION THRESHOLDS FOR DIFFERENT WEIGHTS

| Original Chapter-IV Entities\Weights (300) 1: 15,293 (Chapter-IV); 0: 20,059 (REST) | | |
|---|---|---|
| **Thresholds** | | |
| 1: Sum > 2 | 1: Sum > 3 | 1: Sum > 4 |
| 0: Sum < 2 | 0: Sum < 3 | 0: Sum < 4 |
| 1: 94.43 % | 1: 87.81 % | 1: 78.21 % |
| 0: 15.40 % | 0: 28.05 % | 0: 40.97 % |
| **De-biased Entities\Weights (141) 1: 15,293 (Chapter-IV); 0: 20,059 (REST)** | | |
| **Thresholds** | | |
| 1: Sum > 0.1 | 1: Sum > 0.2 | 1: Sum > 0.3 |
| 0: Sum < 0.1 | 0: Sum < 0.2 | 0: Sum < 0.3 |
| 1: 95.15 % | 1: 89.17 % | 1: 83.41 % |
| 0: 20.61 % | 0: 37.85 % | 0: 50.89 % |
| **Influenced Entities\Weights (146) 1: 15,293 (Chapter-IV); 0: 20,059 (REST)** | | |
| **Thresholds** | | |
| 1: Sum > 0.1 | 1: Sum > 0.2 | 1: Sum > 0.3 |
| 0: Sum < 0.1 | 0: Sum < 0.2 | 0: Sum < 0.3 |
| 1: 95.18 % | 1: 89.74 % | 1: 84.94 % |
| 0: 20.53 % | 0: 36.85 % | 0: 48.51 % |

When compared with the results in Table. 3, the categorization thresholds created using our approach (entities and weights) can outperform the Bi-GRUs + Attention Model by a significant margin in categorizing the short-summaries into Chapter-IV.

### B. Attentional Models for Codes between 280-289

From the list of Chapter-IV ICD codes available in the MIMIC-III dataset, we have selected 3 frequent and 3 non-frequent codes (see Table. 5). Datasets are created for these codes using only the short-summaries that belong to Chapter-IV (15,293). They are labelled based on *one-vs.-rest* approach (see Fig. 4).

**TABLE 5**
6 CHAPTER-IV ICD CODES

| Frequent Codes | Summaries (15,293) | Non-Frequent Codes | Summaries (15,293) |
|---|---|---|---|
| 285.9 | 4387 | 285.22 | 312 |
| 287.5 | 2427 | 288.00 | 217 |
| 286.9 | 791 | 289.9 | 29 |

We have selected the codes that can reflect all the codes that belong to Chapter-IV. There are ICD codes that are very rare (< 30 summaries) in MIMIC-III. To create the attentional models for these codes we should use enough data/summaries to collect the code features, which is not possible with MIMIC-III. It is better not to consider them at all. Researchers have used knowledge bases created using ICD code names and associated synonyms as a workaround for rare labels; however, if we cannot extract a good number of textual features from the summaries, even the KBs won't be enough.

To create the attentional models, we have used two attention mechanisms: dot-product Attention and multi-head Attention. Tabel. 6 showcases the configuration we used for training the models. We have mostly used similar configurations for the codes in the Table. 5. Only the gated recurrent unit dimensions (G) in the BiGRUs model and feed-forward upward projection (Df) in the Transformer model are selected based in the best performance.

**TABLE 6**
CONFIGURATION FOR THE MODELS
(M: MAX. SEQUENCE LENGTH, E: EMBEDDING DIM, G: GRU DIM, D: DROPOUT, B: BATCH SIZE, N: NUMBER OF HEADS, DF: FEED FORWARD UPWARD PROJECTION SIZE, L: LOSS, BCE: BINARY CROSS ENTROPY, SCCE: SPARSE CATEGORICAL CROSS ENTROPY, O: OPTIMIZER)

| Model | Config |
|---|---|
| Bi-GRUs + Attention | M: 512 or 1200; E: 100<br>G: 64 or 128 or 1024<br>D: 0.2; B: 64<br>L: BCE; O: Adam |
| Transformer + Multi-Head Attention | M: 1200; E: 100<br>Df: 64 or 128 or 1024<br>D: 0.2; B: 64<br>L: SCCE; O: Adam |

Most of the existing work with MIMIC-III reported their F1 scores. There are other metrics too; however, due to the imbalanced nature of MIMIC-III, these metrics can be problematic when evaluating the performance of an auto-coding system. F1 scores have been widely used when the data has rare labels [30][31]. F1-score can be interpreted as a weighted average between Precision and Recall, where F1-score reaches its best value at 1 and worst score at 0. The relative contribution of Precision and Recall are equal to the F1-score, and the harmonic mean helps find the best trade-off between the two quantities [ab].

After creating the attentional models for all the codes in Table. 5, we have collected the Micro- and Macro-F1 scores for them (see Table. 7). Averaging the scores allowed to evaluate our chapter-based attention models approach and compare them with existing results. These are some of the results (F1-scores) reported by others using MIMIC-III: 0.48 (Micro) [dd], 0.51 (Mi.) [ee], 0.57 – 0.67 (Mi.) [ff], 0.63 (Mi.) and 0.57 (Macro) [gg]. Comparing these results with F1-scores in Table. 7 showcases that the average values for the Micro and Macro scores from our attentional models are closer to 0.79 – 0.81 and 0.57 – 0.59, suggesting that creating chapter-based attentional models can improve the ICD coding.

**TABLE 7**
MICRO- AND MACRO-AVERAGED F1-SCORES FOR THE 6 CODES
(TR: TRAIN, V: VALID, T: TEST, M1: BI-GRUS + ATTENTION, M2: TRANSFORMER + MULTI-HEAD ATTENTION, EP: EPOCHS, 1: CHAPTER-IV, 0: REST)

| ICD-9 Code | | Model | Micro-F1 | | Macro-F1 | |
|---|---|---|---|---|---|---|
| 285.9 vs Rest | | | M1 | M2 | M1 | M2 |
| 1: 4387 0: 10906 | 1: 4387 0: 1000 | Tr: 3500 V: 500 T: 387 | Ep: 5 0.82 | Ep: 5 0.78 | Ep: 5 0.62 | Ep: 5 0.56 |
| 287.5 vs Rest | | | | | | |
| 1: 2427 0: 12866 | 1: 2427 0: 500 | Tr: 2250 V: 250 T: 427 | 0.84 | 0.85 | 0.57 | 0.50 |
| 286.9 vs Rest | | | | | | |
| 1: 791 0: 14502 | 1: 791 0: 150 | Tr: 720 V: 80 T: 141 | 0.81 | 0.89 | 0.51 | 0.47 |
| 285.22 vs Rest | | | Ep: 10 | | Ep: 10 | |
| 1: 312 0: 14981 | 1: 312 0: 150 | Tr: 375 V: 25 T: 62 | 0.82 | 0.73 | 0.78 | 0.65 |
| 288.00 vs Rest | | | | | | |
| 1: 217 0: 15076 | 1: 217 0: 100 | Tr: 250 V: 25 T: 42 | 0.74 | 0.88 | 0.63 | 0.80 |
| 289.9 vs Rest | | | | | | |
| 1: 29 0: 15264 | 1: 29 0: 15 | Tr: 20 V: 10 T: 14 | 0.71 | 0.71 | 0.42 | 0.42 |
| | | **Average** | **0.79** | **0.81** | **0.59** | **0.57** |

## IV. EXPERIMENTS

### A. Interpretability for NEC Model

The approach we proposed for the NEC model does not require much computation or training. It is a simple approach that uses only 146 entities and their weights to create the categorization thresholds (see Table. 4) that can categorize the short summaries into Chapter-IV or REST. One of the primary and initial steps for the NEC model is using a pre-trained model to extract DISEASE entities [] from the summaries. Using simple regular expression patterns and matching functions, we are able to create weights and find entity matches in the summaries to sum up scores that can fit into the thresholds. This simple approach is as effective as neural networks. However, due to the imbalanced and unstructured nature of the data in MIMIC-III, both approaches can perform incorrect categorizations. One way to tackle this problem is to provide interpretability, which allows us to show the 'human coders' what entities (out of 146) are found in the summaries. These entities can be used to provide information such as anatomical regions, diseases, signs, and symptoms, which can be used for associating with the ICD code names and also provide a way for correcting and validating the categorizations.

We conduct our experiments to identify the thresholds with many incorrect categorizations. Whenever a summary falls into these "faulty" thresholds, a human coder can verify them using the matched entities, as illustrated in Fig. 10. For example, let's consider the thresholds SUM > 0.6 (for Chapter-IV) and SUM < 0.6 (for REST). A total of 25,344 {Chapter-IV: 72%, REST: 71%} summaries form 35,352 short summaries fit those thresholds perfectly (correct categorizations). Table. 8 below lists the thresholds near 0.6 (> and <) that human coders should worry and not worry about. For the thresholds greater than 0.6, **0** indicates incorrect categorizations from the REST summaries that fall into the thresholds in Table. 8. And for thresholds that are less than 0.6, **1** indicates the incorrect categorizations from Chapter-IV summaries.

**TABLE 8**
IDENTIFYING "FAULTY" THRESHOLDS

| Thresholds | Summaries in Chapter-IV (1) and REST (0) | Share in 35,552 |
|---|---|---|
| **0 indicates incorrect categorizations** | | |
| SUM > 0.6 & SUM < 1 | **1: 2419 (43%) 0: 3142 (57%)** | 15% |
| SUM > 1 & SUM < 2 | 1: 6472 (73%) 0: 2429 (27%) | 25% |
| SUM > 1 & SUM < 1.5 | **1: 3661 (67%)**<br>**0: 1783 (33%)** | 15% |
| SUM > 1.5 & SUM < 2 | 1: 2811 (81%)<br>0: 646 (19%) | 10% |
| SUM > 2 & SUM < 3 | 1: 1937 (91%)<br>0: 182 (9%) | 6% |
| **1 indicates incorrect categorizations** | | |
| SUM > 0.3 & SUM < 0.6 | **1: 1945 (30%)**<br>**0: 4567 (70%)** | 18% |
| SUM > 0 & SUM < 0.3 | 1: 2211 (20%)<br>0: 9007 (80%) | 32% |

Here, we calculated the share by dividing the total number of summaries in a threshold (ex: SUM > 0.6 & SUM < 1) with 35,552. Using these values, we identified those faulty thresholds that human coders should focus on. A deeper look into these thresholds can help us identify precise ranges that are problematic. For example, the threshold SUM > 1 & SUM < 2 in Table. 8 is producing 27% incorrect categorization; analyzing it further helped us to identify the precise threshold (SUM > 1 and SUM < 1.5) where incorrect categorizations are more (33%) when compared to 1.5 and 2 (19%). Fig. 10 showcases the entities that belong to a REST summary that has SUM > 1 (1.22). These entities can provide interpretability for human coders.

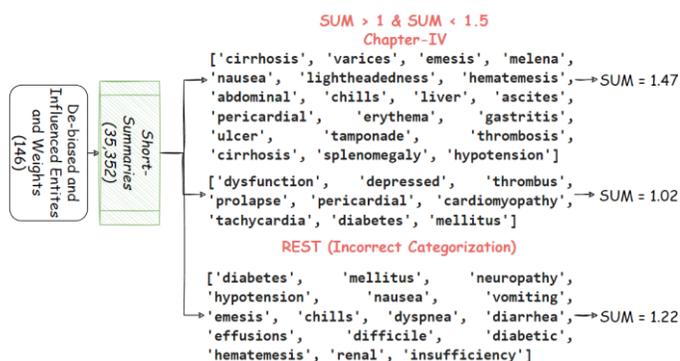

*Figure 10: Interpretability (i.e., Matched Entities) for Incorrect Categorization of REST Summary at SUM > 1 & SUM < 1.5 thresholds*

## V. Conclusion

This study proposes a chapter-based method for improving ICD coding from discharge summaries. Our approach first categorizes the discharge summaries into a chapter, and only then the discharge summaries that belong to that Chapter are used to create data based on the *one-vs-rest* approach. Initially, we processed the full-length discharge summaries by removing unnecessary information/headings that can add noise, thus, allowing us to create short summaries with less vocab size. These short summaries are used in our research to perform chapter-based categorization and create attentional models. We have proposed a simple approach using regular expression patterns and matching functions for the chapter-based categorization to identify/match the important entities from the short summaries. This important entity list is created with the help of a pre-trained NER model that can detect diseases, conditions, signs, and symptoms from clinical texts. We initially extracted the entities from all the summaries using the pre-trained model. We have extracted 375 most common entities from these entities based on document frequency. After applying the de-biasing and influencing criteria, we have decreased the entities to 146 with modified or improved weights. These 146 entities are used to find matches in the short summaries, and the weights are used to sum up, the identified entities. This allows us to create categorization thresholds that fit the data perfectly and provide faulty thresholds and interpretability for the human coders to correct and validate the categorizations.

Several researchers have proposed various methods for automating the ICD coding, but so far, the performance evaluation of these approaches has been limited to carefully curated and clean benchmark datasets (MIMIC-III), which may not be reflective of how well these systems perform in the real world. As a whole, MIMIC-III only includes around 9,000 ICD-9 codes, most of which are unbalanced. The MIMIC-III results from existing methods show that Micro-average F1-scores are between 0.4 and 0.7, indicating that there are still many false positives. Our approach achieved average Micro-F1 Scores (6 codes in Table. 7) of 0.79 and 0.81 for our models (BiGRU, Transformer), showcasing significant improvements in the performance of ICD coding.